\documentclass[10pt,conference,letterpaper]{IEEEtran}
\IEEEoverridecommandlockouts
\usepackage{diagbox}
\usepackage{balance}
\usepackage{subfig}
\usepackage{booktabs}
\usepackage{cite}
\usepackage{amsmath,amssymb,amsfonts}
\usepackage{algorithm}
\usepackage{algpseudocode}
\usepackage{graphicx}
\usepackage{textcomp}
\usepackage{xcolor}
\usepackage{caption}
\usepackage{multicol}
\usepackage{multirow}
\usepackage{makecell}
\usepackage{cleveref}
\def\BibTeX{{\rm B\kern-.05em{\sc i\kern-.025em b}\kern-.08em
    T\kern-.1667em\lower.7ex\hbox{E}\kern-.125emX}}
\usepackage{fancyhdr}

\begin{document}

\title{Interpretable Modeling of Deep Reinforcement Learning Driven Scheduling}

\author{\IEEEauthorblockN{Boyang Li}
\IEEEauthorblockA{\textit{Department of Computer Science} \\
\textit{Illinois Institute of Technology}\\
Chicago, USA \\
bli70@hawk.iit.edu}
\and

\IEEEauthorblockN{ Zhiling Lan}
\IEEEauthorblockA{\textit{Department of Computer Science} \\
\textit{University of Illinois Chicago}\\
Chicago, USA \\
zlan@uic.edu}
\and
\IEEEauthorblockN{Michael E. Papka}
\IEEEauthorblockA{
\textit{Argonne National Laboratory}\\
\textit{University of Illinois Chicago} \\
Lemont, USA \\
papka@anl.gov}
}

\maketitle

\thispagestyle{fancy}
\lhead{}
\rhead{}
\chead{}
\lfoot{\footnotesize{
MASCOTS2023, October 16-18,2023, Stony Brook, NY, USA
\newline 979-8-3503-1948-4/23/\$31.00 \copyright 2023 IEEE}}
\rfoot{}
\cfoot{}
\renewcommand{\headrulewidth}{0pt}
\renewcommand{\footrulewidth}{0pt}

\begin{abstract}
In the field of high-performance computing (HPC), there has been recent exploration into the use of deep reinforcement learning for cluster scheduling (DRL scheduling), which has demonstrated promising outcomes. However, a significant challenge arises from the lack of interpretability in deep neural networks (DNN), rendering them as black-box models to system managers. This lack of model interpretability hinders the practical deployment of DRL scheduling.  
In this work, we present a framework called IRL (Interpretable Reinforcement Learning) to address the issue of interpretability of DRL scheduling. The core idea is to  interpret DNN (i.e., the DRL policy) as a decision tree by utilizing imitation learning. Unlike DNN, decision tree models are non-parametric and easily comprehensible to humans.
To extract an effective and efficient  decision tree, IRL incorporates the Dataset Aggregation (DAgger) algorithm and introduces the notion of critical state to prune the derived decision tree. Through trace-based experiments, we demonstrate that IRL is capable of converting a black-box DNN policy into an interpretable rule-based decision tree while maintaining comparable scheduling performance. Additionally, IRL can contribute to the setting of rewards in DRL scheduling.

\end{abstract}
\begin{IEEEkeywords}
cluster scheduling; deep reinforcement learning; high-performance computing; interpretation; decision tree
\end{IEEEkeywords}

\section{Introduction} \label{sec:intro}
Cluster scheduling, also known as batch scheduling,  is pivotal to high-performance computing (HPC). It is responsible  for determining the order in which jobs are executed on an HPC system. Heuristics play a significant role in cluster scheduling, with the first-come, first-served (FCFS) policy being a widely employed scheduling approach on production systems \cite{mu2001utilization}.   To improve system utilization,  backfilling is  commonly used in cluster scheduling to enhance system utilization, which allows subsequent jobs to be moved ahead to utilize available resources \cite{mu2001utilization}.

Reinforcement learning,  a subfield of machine learning,  focuses on the automatic learning of decision-making strategies to maximize cumulative rewards through interactions with the environment\cite{sutton2018reinforcement}. 
Deep reinforcement learning (DRL), which combines reinforcement learning with deep neural networks, has been employed for cluster scheduling \cite{mao2016resource,li2022encoding, decima, fan2022dras}, demonstrating promising results. Unfortunately, very few, if any, of these approaches have been deployed on real-world production systems.
One key hurdle is \emph{the lack of model interpretability}. The superior performance of DRL scheduling stems from its deep neural network (DNN) \cite{lecun2015deep}; however, DNN appears as a black-box to system managers \cite{dethise2019cracking},  making it challenging to comprehend, debug, deploy, and adjust in practice. As a result, system managers have reservations about using DRL scheduling on production systems.  Therefore, it is essential to develop interpretable models that facilitate the practical deployment of DRL scheduling. 

Many techniques have been developed to understand the behaviors of DNNs; however, there are two issues when applying these techniques to interpret DRL scheduling. First, existing techniques predominantly focus on monitoring neuron activations to identify the features that trigger them \cite{bau2017network}. 
Consequently, system managers still need to possess knowledge of machine learning. Second, the current DNN interpretation methods are primarily designed for well-structured vector inputs such as images \cite{du2018towards,wang2019deepvid} and sentences \cite{liu2021deep, toneva2019interpreting}, which are not applicable to the cluster scheduling problem. 
We believe there is a pressing need to provide a simple, deterministic, and easily understandable model for interpreting DRL scheduling.

Decision tree is non-parametric, and easy for humans to understand. %They can represent complex policies.
However, training a decision tree in the context of  DRL poses significant challenges\cite{viper2018}. While some studies have attempted to train some decision tree policies for reinforcement learning  \cite{ernst2005tree}, subsequent work pointed out that these approaches struggled with relatively simple problems such as cart-pole \cite{viper2018}. Other efforts have sought to convert the DNNs employed in DRL to decision trees \cite{viper2018,frosst2017distilling}. This conversion is built
upon a teacher-student training process, where the DNN policy serves
as the teacher, generating input-output samples to construct
a student decision tree for classification or regression.  Compared to directly training decision tree policies for reinforcement learning, these converted decision trees can achieve better performance and handle more complex problems \cite{viper2018, frosst2017distilling}.

In this work, we present IRL, an \textbf{I}nterpretable  modeling for deep \textbf{R}einforcement \textbf{L}earning scheduling. IRL is based on imitation learning \cite{hussein2017imitation}. In simple terms, a decision tree is trained to mimic the DRL agent (i.e., deep neural network). 
Specifically, once a DRL agent (deep neural network) is built for cluster scheduling, this agent is used to generate input-output samples. These samples are used to train a decision tree policy.  However, we observe that the decision tree obtained through this straightforward conversion process does not closely resemble the original DRL agent as expected.

To overcome this issue, we employ the DAgger algorithm \cite{ross2011reduction} to improve the scheduling performance of the decision tree. 
Furthermore, we have found that existing conversion algorithms\cite{loh2011classification} tend to produce a large decision tree with an excessive number of branches, which will greatly impact the effectiveness of the decision tree for decision making. To reduce the size of the decision tree, we introduce the concept of critical state in the  scheduling environment. A critical state is defined as a system state that has a nontrivial impact on scheduling performance. %It is motivated by the observation that the job selection in a relatively idle system state has minimal impact on the scheduling performance. 
By utilizing the critical state, we can decrease the size of the decision tree while still maintaining effectiveness.
%Through the use of critical state, we can reduce the size of the decision tree  (i.e., higher efficiency) while maintaining effectiveness. 
Specifically, we make three major contributions in this work:
\begin{itemize}
  \item We present IRL, an interpretable modeling for DRL scheduling. IRL  converts a black-box DRL policy to an easy-to-understand decision tree policy, thereby overcoming the lack of interpretability issue of DRL scheduling. 
  \item We showcase how the interpretability of IRL can assist in the design of DRL scheduling, such as reward setting.
  \item  Our trace-based experiments demonstrate the decision tree derived from IRL achieves comparable scheduling performance to the original DRL scheduling. In contrast to the black-box deep neural network, the decision tree is straightforward to comprehend, debug, and modify.

\end{itemize}

\section{Background} \label{sec:background}

\subsection{Cluster Scheduling in HPC} \label{secc:HPC sch}
A cluster scheduler is responsible for allocating resources and for determining the order in which jobs are executed on an HPC system. When submitting a job, a user is required to provide two pieces of information: the number of compute nodes required for the job (i.e., job size) and job runtime estimate (i.e., walltime). The scheduler
determines when and where to execute the job. The jobs are stored and sorted in the waiting queue based on a site's policy. The scheduler determines \textit{when and where} to execute the jobs \cite{li2019effect}. Once a new job is submitted, the job scheduler sorts all the jobs in the waiting queue based on a job prioritizing policy. A number of popular job
prioritizing policies have been proposed, and one of the widely used policies is FCFS \cite{mu2001utilization}, which sorts jobs in the order of job arrivals. 

In addition, backfilling is a commonly used approach to enhance job scheduling by improving system utilization, where subsequent jobs are moved ahead to utilize free resources. A widely used strategy is EASY backfilling  which allows short jobs to skip ahead under the condition that they do not delay the job at the head of the queue \cite{mu2001utilization}. 

%HPC jobs differ from data center workloads, such as those at Google and Microsoft. On sites like Google's and Microsoft's, there are various types of workloads, including long-running services and directed acyclic graph (DAG) jobs. Each job comprises several tasks, and the resource requirements for each task can sometimes be malleable. The typical minimum scheduling unit is a task. In contrast, HPC was traditionally dominated by single, rigid jobs that couldn't be decomposed. Here, the minimum scheduling unit is usually an entire job.
 While cluster scheduling is an active area of research in both HPC and cloud computing, these two communities target different workloads, resulting in divergent research approaches.
In data centers such as those at Google or Microsoft, typical workloads include long-running services and directed acyclic graph (DAG) jobs. Each job comprises multiple tasks, and the resource requirements for these tasks are often adjustable (i.e., malleability). In this context, the smallest scheduling unit is typically a task.
In contrast, HPC is characterized by tightly-coupled parallel jobs (i.e.,  rigidity). Here, the scheduling unit is usually an entire job.

\begin{table*}
\begin{center}
 \caption{Comparison of heuristics, DRL scheduling, and IRL (Interpretable Reinforcement Learning).}

  \label{tab:comparison of scheduling policy}
 \begin{tabular}{|l|c|c|c|c|}
\hline
\diagbox[width = 16em]{\thead{Features}}{\thead{Methods}}& \thead{Heuristics \\
\cite{mu2001utilization} }& \thead{ DRL scheduling \\ \cite{fan2022dras,li2022encoding,mao2016resource}}& \thead{IRL} \\
\hline
\thead{Not black-box }&$\sqrt{}$  & $\times$&$\sqrt{}$   \\
\hline
\thead{Easy to comprehend}& $\sqrt{}$  & $\times$&$\sqrt{}$  \\

\hline
 \thead{Superior scheduling performance}&$\times$ & $\sqrt{}$ & $\sqrt{}$   \\
 \hline
\thead{Rapid decision making}& $\sqrt{}$  & $\times$&$\sqrt{}$  \\
\hline
\end{tabular}
\end{center}
\vspace{-0.5cm}
\end{table*}
\subsection{Deep Reinforcement Learning Scheduling} \label{sec:DFP overreview}
\begin{figure}
\centering
\includegraphics[height=1.3in,width=3in]{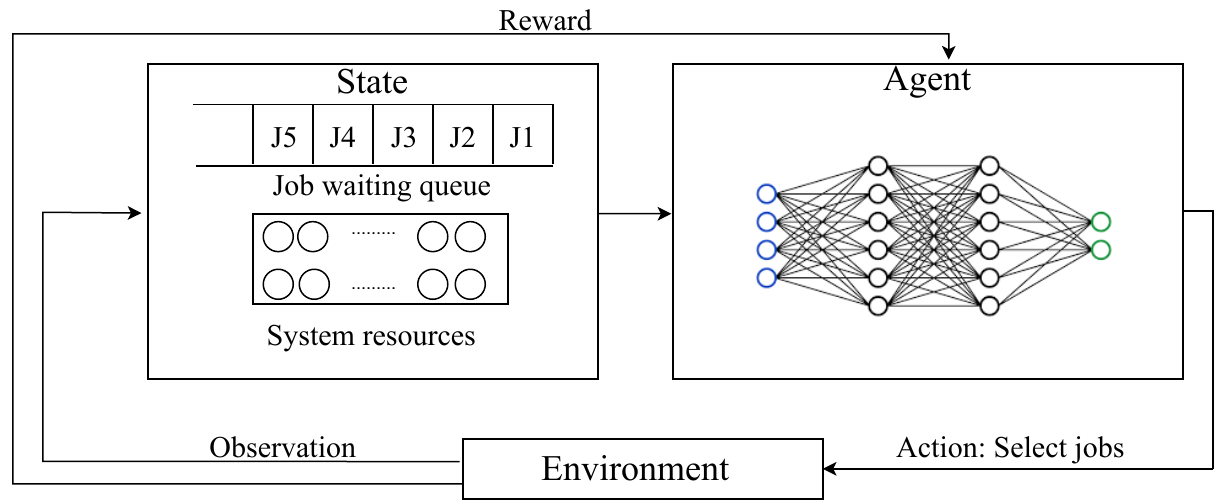}

\caption{The interaction between environment and an agent in reinforcement learning.}
\label{fig:rl}
\vspace{-0.4cm}
\end{figure}

Reinforcement learning is an area of machine learning that is primarily focused on dynamic decision making where an intelligent agent takes actions in an environment
with the goal of maximizing some reward \cite{sutton2018reinforcement}. The recent advancement of reinforcement learning enhanced with deep neural networks has yielded a number of promising performances for cluster scheduling \cite{mao2016resource,li2022encoding, decima, fan2022dras}. In DRL driven scheduling, the agent is trained to learn  a proper scheduling policy according to a specific scheduling objective (e.g., reward) provided by system managers. Once trained, the agent can automatically interact with the scheduling environment and dynamically adjust its policy as workload changes.  Since the state space is typically enormous, memorizing all states  becomes  infeasible. DRL driven scheduling uses a deep neural network for approximation \cite{mnih2013playing}. Figure \ref{fig:rl} shows an overview of typical DRL driven scheduling. At each step, state is observed and fed to the scheduling agent. The agent provides job selection and receives the reward as the feedback.

\subsection{Related Work} \label{sec:related work}

Interpreting deep neural networks is an active topic in the machine learning area \cite{du2018towards,liu2021deep,feng2020visualizing, wu2022layer}. The interpretation methods can be broadly classified into two categories \cite{molnar2020interpretable}. One is \emph{local methods}, such as LIME \cite{ribeiro2016should}. These methods focus on explaining the prediction of a single instance in the dataset. The other is \emph{global methods} \cite{du2019techniques}, which aim to explain the average behavior of a machine learning model. In this work, we adopt global methods because our objective is to interpret the average behavior of a DRL-driven scheduling model. 

Many global interpretability approaches focus on understanding the mechanism of DNNs, such as convolutional neural networks (CNN) \cite{du2018towards,liu2021deep} and recurrent neural networks (RNN) \cite{feng2020visualizing, wu2022layer}. These approaches still require the system manager to have knowledge of machine learning, which is not the goal of IRL. In addition, these approaches are designed for well-structured inputs such as images and sentences, and thus are not suitable for deep reinforcement learning driven scheduling problems.

Another way to achieve interpretability is to use only a subset of algorithms that create interpretable models. A decision tree is a commonly used interpretable model \cite{kotsiantis2013decision}. A decision tree is a decision-support hierarchical model that uses a tree-like model of decisions and their possible consequences. It can represent a complex policy. However, the main hurdle is that the decision tree is hard to train directly in reinforcement learning problems. There have been work training decision tree policies for reinforcement learning \cite{ernst2005tree}, but the following research pointed out that this method could not even achieve satisfactory performance in a not complicated reinforcement learning problem such as cart-pole \cite{viper2018}. To overcome this obstacle,  some researchers refer to the idea of imitation learning to convert the DNN to a decision tree \cite{frosst2017distilling, viper2018}. The conversion is built on top of a teacher-student training process, where the DNN policy acts as the teacher and generates input-output samples to construct the student decision tree. Compared to directly training decision tree policies for reinforcement learning, the converted decision tree achieves better performance in many reinforcement learning problems \cite{frosst2017distilling,viper2018}.

Interpreting DNNs to decision trees has shown promising results in many areas \cite{meng2020interpreting,hu2022primo,schmidt2021can}. Meng et al. presented Metis to interpret deep learning based networking systems based on decision trees and hypergraphs \cite{meng2020interpreting}. Hu et al. demonstrated the decision tree converted from DRL achieved satisfactory performance in learning adaptive bitrate (ABR) algorithms \cite{hu2022primo}. Schmidt et al. applied the conversion from DNN to the decision tree in autonomous driving \cite{schmidt2021can}. However, there is no effort made to interpret the DNNs in DRL driven scheduling in HPC. \emph{To the best of our knowledge, IRL is the first attempt to interpret DRL scheduling in HPC}. Table \ref{tab:comparison of scheduling policy} summarizes the key features of heuristics, DRL scheduling, and the proposed IRL --- interpretable reinforcement learning scheduling.

\begin{figure*}
\centering
\includegraphics[height=1.5in,width=7in]{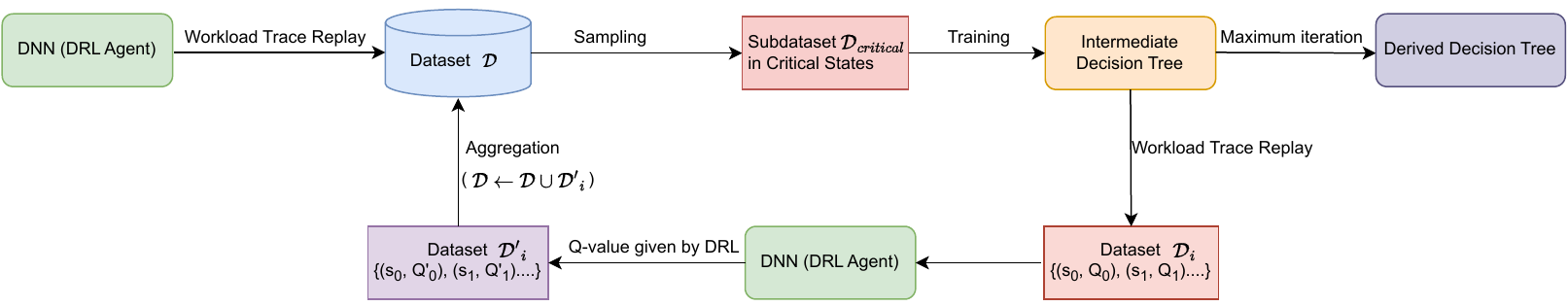}

\caption{Overview of IRL design. A cylinder represents a data repository $\mathcal{D}$. A rounded rectangle denotes a scheduling policy, which is either DRL or a decision tree. %Rounded rectangles with the same color represent the same scheduling policy.
A rectangle represents a datatset. It is  sampled from $\mathcal{D}$, or produced by a scheduling policy (decision tree or DRL).   }
\vspace{-0.2cm}
\label{fig:IRL}
\end{figure*}

\section{IRL Design} \label{sec:DFPSch Design}

IRL, shown in Figure \ref{fig:IRL},  is developed to provide an interpretable model for general DRL scheduling in HPC. The design of IRL  is based on imitation learning, where the DNN policy of the DRL agent acts as the teacher and generates input-output samples to construct the student decision tree. Specifically, a trained deep neural network is obtained by training a workload trace. This neural network acts as the teacher and generate input-output samples from the original training workload trace. These generated input-output samples are used as the training dataset to train an interpretable decision tree.

There are two issues in the above process: (1) the derived decision tree might not resemble the original deep neural network very well, and (2) the size of the decision tree could be huge. To overcome these obstacles, we employ two techniques. We integrate the DAgger algorithm \cite{ross2011reduction} to address the former issue, and then introduce the critical state concept for the latter one. 

The details are described in the subsequent subsections. In order to make the description of our method more straightforward, we use deep Q-network (DQN) as an illustrative example. In Section \ref{sec:network policy}, we describe DQN driven scheduling. In Section \ref{sec:dt conversion}, we present the conversion from the DQN policy to the decision tree policy.  We describe the generation of the input-output sample datatset by DQN, and the generation of the decision tree with the sample dataset. We also describe how to integrate DAgger in our design to improve the scheduling performance of the decision tree, and how to reduce the size of the decision tree via the concept of critical state.

\subsection{DQN  Scheduling} \label{sec:network policy}
Recent studies have employed various deep reinforcement learning (DRL) methods for cluster scheduling \cite{decima, fan2022dras}. 
Regardless of the specific DRL method employed, the underlying principle remains consistent. In this study, we utilize DQN \cite{mnih2013playing} as a practical example. The scheduling system, driven by a DRL agent, follows the approach outlined in \cite{fan2022dras}. The DRL scheduling agent aims to optimize scheduling performance by making decisions on when and which jobs in the waiting queue should be allocated to available computer resources. At each scheduling instance, the agent encodes both job and system information into a vector, which is fed as an input to the neural network. Based on the neural network's output, the agent selects jobs from the wait queue and then receives a reward signal from the scheduling environment.

In DQN, the deep neural network is utilized to approximate Q-value as $Q(s_{k}, a_{k})$ (i.e., the expected cumulative reward of
taking an action $a_{k}$ in the state of $s_{k}$). DQN processes
one job at a time and produces the expected Q-value for this
job. The input of DQN is a 1-D vector containing job size (i.e., number of nodes requested), job length (i.e., estimated job runtime), and system utilization. The output is a single neuron corresponding to the expected Q-value of the
job. The scheduler enforces a window of jobs at the front of the job wait queue. The window alleviates the job starvation problem by providing higher priorities to older jobs. 
%We use the same network to approximate Q-value for all the jobs in the window.

In order to explore various actions,  during the training, the agent follows the $\epsilon$-greedy policy (i.e., the agent randomly chooses a job instead of the job with the highest Q-value with probability $\epsilon$). During the testing or inference time, the agent selects the job with the highest Q-value.

\subsection{Decision Tree Conversion} \label{sec:dt conversion}
Decision tree is a supervised learning method commonly used for classification or regression. In this work, we use decision tree for regression. 
Specifically, in order to interpret the DQN policy, our IRL works as follows. The input of the decision tree is the state, and the output of the decision tree is the imitated output (Q-value) of DQN. 
The DQN scheduling agent replays the workload trace to produce a trajectory of (state, Q-value) pairs. This trajectory will be used as the training dataset $\mathcal{D}$ of the decision tree so that the output of the decision tree will approximate the output of the DQN. 

The process of extracting an effective and efficient decision tree to interpret the DQN policy presents several challenges. Firstly, we observed that the decision tree trained once may not resemble the DQN policy well. When replaying the same workload trace, the decision tree agent may choose a different job from the DQN agent, since the imitated Q-value by the decision tree is unlikely to precisely match the Q-value output by DQN.  Consequently, due to the varying job selection, the decision tree is prone to jump to a state which it has never seen in the training dataset $\mathcal{D}$, and behaves unpredictably in these unseen states. 

To address this issue, we incorporate DAgger \cite{ross2011reduction} into the decision tree conversion. DAgger is an iterative training algorithm. Instead of training the decision tree one time, we train it multiple times. After each training iteration, the newly generated decision tree is used to replay the workload trace and  demonstrate its policy to the DQN agent. As a result, a new trajectory of (state, Q-value) pairs following the newly generated decision tree policy is produced, denoted as $\mathcal{D}_{i}$. This new trajectory may contain unseen states in the training dataset $\mathcal{D}$. As a teacher, the DQN agent assigns the Q-values to the states in this new trajectory, and aggregate this newly produced trajectory $\mathcal{D'}_{i}$ into the training dataset $\mathcal{D}$. The updated dataset $\mathcal{D}$ is then used to train the decision tree in the next iteration. This process repeats multiple times until the maximum iteration is reached. In this work, the maximum iteration is set to 5. The middle part of Figure \ref{fig:IRL} illustrates the iterative process of generating the decision tree.

Next, another challenge of the decision tree conversion is that the generated decision tree with DAgger is normally huge\cite{viper2018}. The cost of decision time is proportional to the size of the decision tree. Hence a smaller-sized decision tree without sacrificing scheduling performance is desired. To overcome the issue, IRL introduces the concept of \emph{critical state} with the goal of generating a reduced-sized decision tree.

Our design is based on a key observation, that is,  when the system is relatively idle (i.e., few jobs are in the waiting queue), the job selection has less impact on the scheduling performance. There are two reasons. First, the range of job selection is limited when there are a few jobs in the waiting queue. For instance, when there is only one job in the waiting queue, this job is selected anyway. 
Similarly, if the available resources can accommodate only one job, it is chosen among all the waiting jobs. 
Second, even if a job is not selected at this scheduling instance, it is likely to be scheduled shortly thereafter.

In the design of IRL, we define \textit{critical state} as the system state when the number of jobs in the waiting queue is greater than a threshold, and \textit{non-critical state} as the state when the number of jobs in the waiting queue is less than or equal to the threshold. 
%If the minimum  number of waiting jobs is set to a large value to define as critical state, the performance of the tree is likely to decrease. 
The threshold should be chosen based on the workload to balance the tree size and the scheduling performance.
%In this study, we set the threshold to three  to balance the tree size and its performance.  
Only samples with  critical states ($\mathcal{D}_{critical}$) in the dataset $\mathcal{D}$ are used to generate the decision tree. Put together, Algorithm 1 shows the complete pseudo code of the IRL method.

\begin{algorithm}
\caption{IRL algorithm}

\begin{algorithmic}[1]

    \State Initialize DataSet $\mathcal{D}$ $\leftarrow$ Trajectory (state, Q-value) generated by DQN via workload trace replay 
    \For{i = 1 to N}
        \State $\mathcal{D}_{critical}$ $\leftarrow$ $\mathcal{D}$ in critical states
        
        \State Train decision tree $DT_{i}$ from $\mathcal{D}_{critical}$
       
        \State DataSet $\mathcal{D}_{i}$ $\leftarrow$ Trajectory generated by $DT_{i}$  via workload trace replay
        \State $\mathcal{D'}_{i}$ $\leftarrow$ Q-values are assigned by DQN for the states in $\mathcal{D}_{i}$ 
        \State $\mathcal{D}$ $\leftarrow$ $\mathcal{D}$ $\cup$ $\mathcal{D'}_{i}$
        
    \EndFor

\end{algorithmic}
\end{algorithm}

\section{Evaluation} \label{sec:eval}

We have implemented IRL using TensorFlow \cite{IRL}. Our deployment of IRL involves integrating it with the discrete-event driven scheduling simulator CQSim \cite{CQSim}. To thoroughly evaluate the performance of IRL, we conduct extensive trace-based simulations using actual workload traces.

In this section, we describe our evaluation methodology, and the experimental results are listed in the next section.

\subsection{Workload Traces} \label{sec:trace}
%We implement IRL with TensorFlow \cite{IRL} in the discrete-event driven scheduling simulator CQSim \cite{CQSim}. The enhanced CQSim is used in the evaluation. 
In our evaluation, two real workload traces are used \cite{feitelson2014experience,PWA}. Table \ref{tab:workload}
summarizes the two workload traces. 
For each trace,  10,000 jobs are employed for training to build both the DQN policy and the decision tree. The convergence of the DQN policy is confirmed by assessing its convergence rate. Subsequently, an additional set of 2,500 unseen jobs is utilized for inference testing.

\subsection{Comparison Methods} \label{comparison method}
We compare IRL with the following three methods.
\begin{itemize}
    \item \textbf{FCFS} represents first come first served method, which is
the default scheduling policy deployed on many production supercomputers \cite{mu2001utilization}. FCFS prioritizes jobs based on their arrival times.
  \item \textbf{DQN} denotes the reinforcement learning scheduling policy \cite{fan2022dras}. It acts as the teacher to construct the decision tree policy.  
  \item \textbf{DAgger} represents a generated decision tree policy without the introduction of critical state.
\end{itemize}

In addition, backfilling is adopted in each of these methods to mitigate resource fragmentation \cite{mu2001utilization}. FCFS comes with EASY Backfilling \cite{mu2001utilization}. In DQN and IRL, the backfilled job is selected by the agent.

\subsection{Experiment Setup} \label{sec:setup}
 
When training the DQN agent, the reward is set to $\sum_{j\in J}-\frac{1}{t_j}$ \cite{mao2016resource}, 
 where $J$ is the set of jobs currently in the system, $t_j$ is the (ideal) running time of the job. This reward function aims to minimize the average job slowdown. The window size of the waiting queue is set to 20. $\epsilon$ is set to 1.0 at the beginning of
the training and decays at the rate of $\alpha$ = 0.995. The input layer contains three neurons, and three fully-connected hidden layers activated by rectified linear units (ReLU) \cite{agarap2018deep} with 32, 16, 8 neurons are used separately. 
The output layer contains one neuron. 

In our experiments, we set the critical state threshold to three, meaning a critical state is defined when the number of jobs in the waiting queue is more than three. Our sensitivity study indicates that for these workloads, this configuration can  balance the tree size and its performance. 
Scikit-learn library is used to generate the decision tree \cite{pedregosa2011scikit}.

\subsection{Evaluation Metrics} \label{sec:metrics}
%Since the objective of DQN is set to minimize the average job slowdown, which is a user-level metric. 
Following the common practice\cite{feitelson1998metrics}, we use the following metrics to evaluate different scheduling methods. 
\begin{enumerate}

 \item \emph{Average job wait time}: the average interval between job submission to job start time. 
 
 \item \emph{Average job slowdown}: the average ratio of job response time (job runtime plus wait time) to the actual runtime, representing the responsiveness of a system.
\end{enumerate}

\begin{table}
\centering
\caption{Workload Traces}
  \label{tab:workload}

\begin{tabular}{cccc}
    \toprule
  Workload &Site &System Size & Period\\
    \midrule
    
SP2& SDSC & 128 & April, 1998-April, 2000\\
DataStar& SDSC &1,664 & March, 2004-April, 2005 \\
     \bottomrule
\end{tabular}
\end{table}

\section{Results} \label{sec:result}
In this section, we present the experimental results. Our analysis centers upon four questions: 

\begin{enumerate}

\item What can IRL contribute to DRL driven scheduling? (Section \ref{sec:reward setting}.)

\item Does the decision tree obtained through IRL exhibit comparable scheduling performance to DRL scheduling?  (Section \ref{sec:scheudling performance})

\item Does the use of critical state reduce the size of the decision tree? (Section \ref{sec:dt_size})

\item Does IRL introduce less runtime overhead than DRL scheduling? (Section \ref{sec:overhead})
\end{enumerate}

\subsection{Reward Setting} \label{sec:reward setting}
Neural networks used in DRL function as black-box models. However, since the decision tree obtained through IRL imitates DRL, we can indirectly understand the DRL scheduling policy by examining the resulting decision tree.

In order to illustrate the use of IRL, we conduct a case study to demonstrate the contribution of IRL to DRL scheduling in terms of reward setting. Specifically, we utilize the SP2 workload as our experimental scenario. Within this case study, we explore two different reward settings: one that aligns with our scheduling objective, and the other that fails to do so. Through this use case, we showcase the interpretability of IRL in identifying key features that significantly influence the decision-making process of the DRL agent. Furthermore, IRL enables us to explain the rationale behind whether a particular reward setting can achieve the scheduling objective or not.

In this case study, the objective of the DQN agent is to minimize the average job slowdown. Two reward settings are used: (1) \emph{Reward A} --- the reward is set to $\sum_{j\in J}-\frac{1}{t_j}$ \cite{mao2016resource}, and (2) \emph{Reward I} --- the reward is set to $\sum_{j\in J}-\frac{(w_j+t_j)}{t_j}$, where $J$ is the set of jobs currently in the system, $t_j$ is the (ideal) running time of the job, and $w_j$ is the waiting time of the job. 
Reward I appears reasonable since it aims to maximize the negative values of all the jobs' slowdown in the system, and our objective is to minimize the average job slowdown. Since job waiting time is in the reward, we also add job waiting time in the input feature. 
%A DQN agent is obtained after training with this inappropriate reward. 

\begin{figure}
\centering
\includegraphics[height=1.7in,width=3.6in]{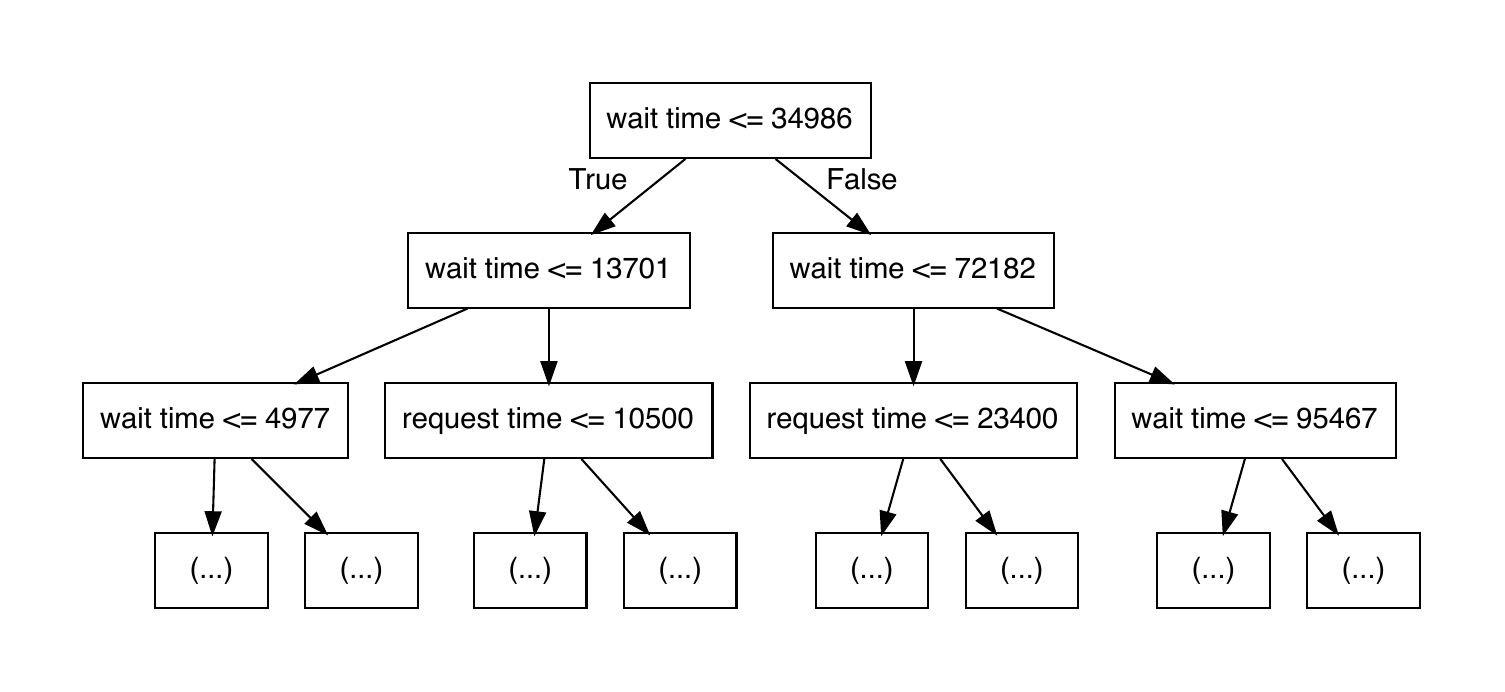}

\caption{Decision tree (depth=10) generated by IRL from the DQN agent with \emph{Reward I}. 
%The depth of the decision tree is set to ten. 
Only the first two depths are presented in the figure due to the space limitation. 
Note that the decision tree's branches primarily revolve around job wait time for decision-making.}
\vspace{-0.2cm}
\label{fig:dt_wrong}
\end{figure}
Figure \ref{fig:dt_wrong} shows the decision tree generated by IRL from this DQN agent using \emph{Reward I}. The depth of the decision tree is set to ten, and only the first two depths are presented due to space limitation. It can be observed that this decision tree primarily bases its decisions on job waiting time. Further analysis reveals that this decision tree favors jobs with longer wait times, and its approach closely aligns with FCFS. The underlying reason is that if we want to maximize  $\sum_{j\in J}-\frac{(w_j+t_j)}{t_j}$, the DQN agent is trained to select (remove) the job with the minimum value of $-\frac{(w_j+t_j)}{t_j}$. A longer wait time leads to a smaller value of $-\frac{(w_j+t_j)}{t_j}$. In the end, the DQN is trained to prefer early coming jobs. If the scheduling objective is to minimize average slowdown, each scheduling decision is supposed to consider the future job wait time instead of the past job wait time. Therefore, the past job wait time should not be a factor in either the reward or input feature.

\begin{figure}
\centering
\includegraphics[height=1.7in,width=3.6in]{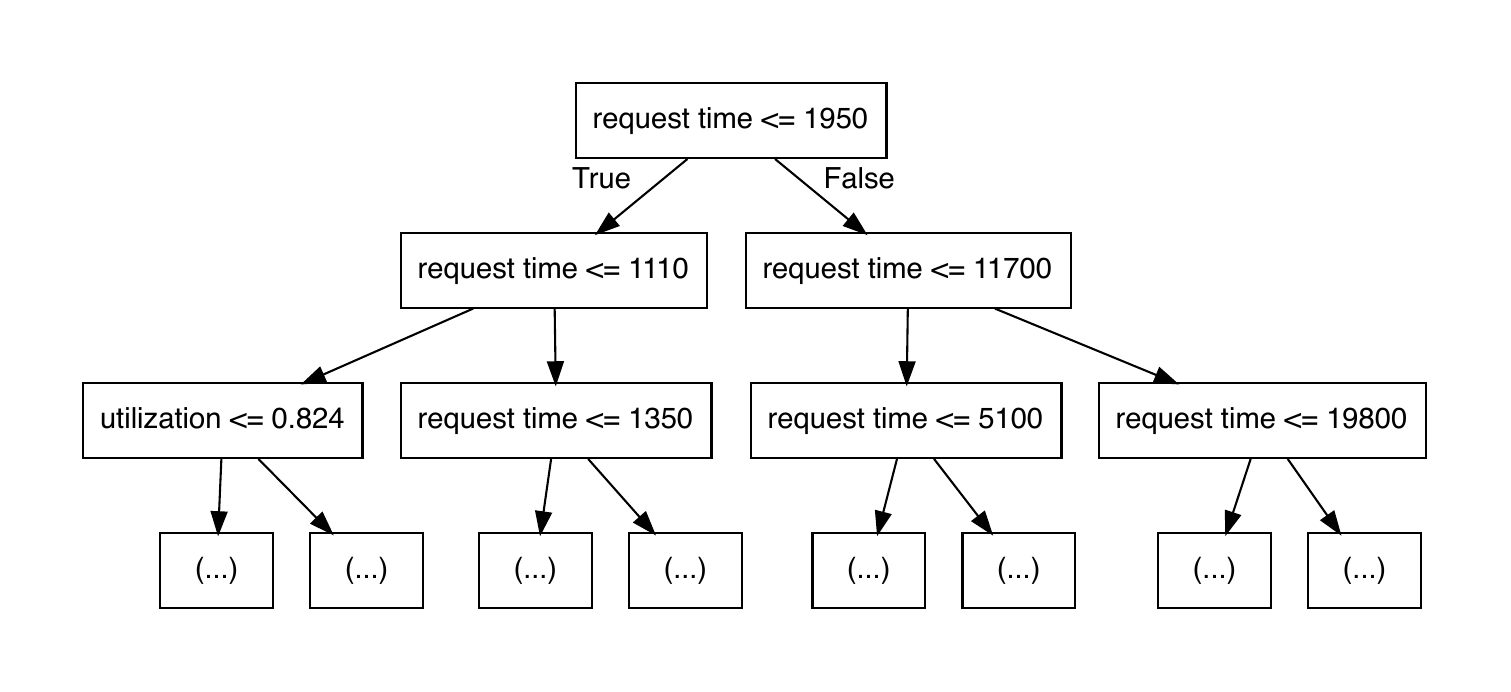}

\caption{Decision tree (depth=10) generated by IRL from the DQN agent with \emph{Reward A}. 
%The depth of the decision tree is set to ten. 
Only the first two depths are presented in the figure due to the space limitation. Note that the decision tree's branches mainly revolve around requested running time for decision-making. }
\vspace{-0.2cm}
\label{fig:dt_right}
\end{figure}

Figure \ref{fig:dt_right} shows the decision tree generated by IRL from the DQN agent using \emph{Reward A}. It can be observed that this decision tree primarily bases its decisions on the job length (the requested running time). Further analysis finds that the decision tree prefers to select short jobs. Indeed, according to the definition of job slowdown, the scheduling objective of minimizing average job slowdown tends to allow shorter jobs to wait less time. This demonstrates that the interpreted DQN policy indicates that the reward setting -- \emph{Reward A} in the DQN is appropriate.

\begin{figure} 
    \centering
  \subfloat[Average job wait time]{%
       \includegraphics[width=0.50\linewidth]{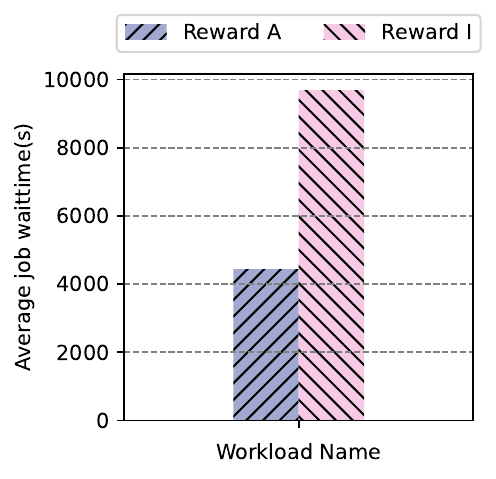}}
    \label{2a}\hfill
  \subfloat[Average job slowdown]{%
        \includegraphics[width=0.47\linewidth]{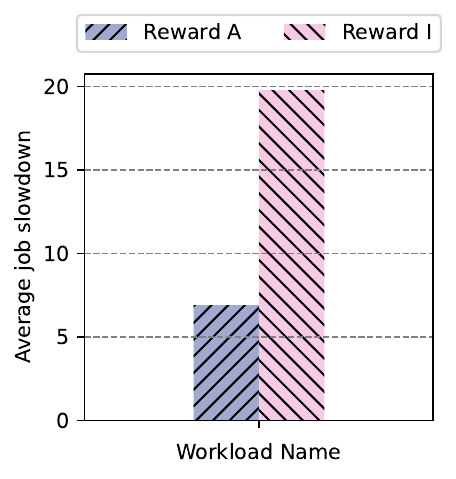}}
  \caption{ Comparison of scheduling performance with DQN  under different reward settings.}
  \label{fig:right_wrong} 
 \vspace{-0.3cm}
\end{figure}

Figure \ref{fig:right_wrong} shows the DQN scheduling performance under different reward settings. Appropriate reward setting (Reward A) can reduce the average job wait time and slowdown by up to 66\% compared to inappropriate reward setting. This study clearly illustrates that IRL contributes to analyzing reward settings in DRL driven scheduling.

\vspace{-0.1cm}

\begin{figure*} 
    \centering
  \subfloat[Average job wait time]{%
       \includegraphics[width=0.40\linewidth]{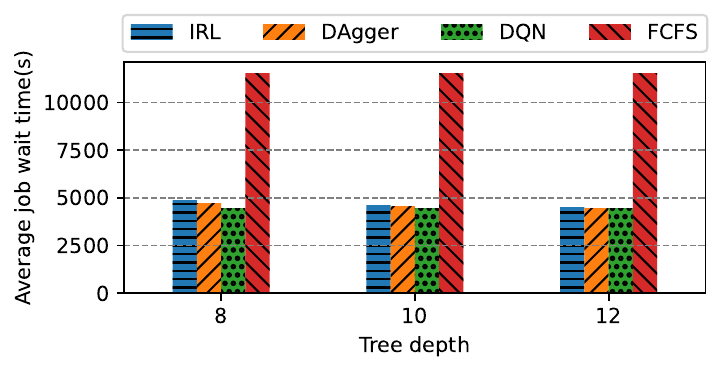}}
    \label{3a}\hspace*{0.06\textwidth}
  \subfloat[Average job slowdown]{%
        \includegraphics[width=0.40\linewidth]{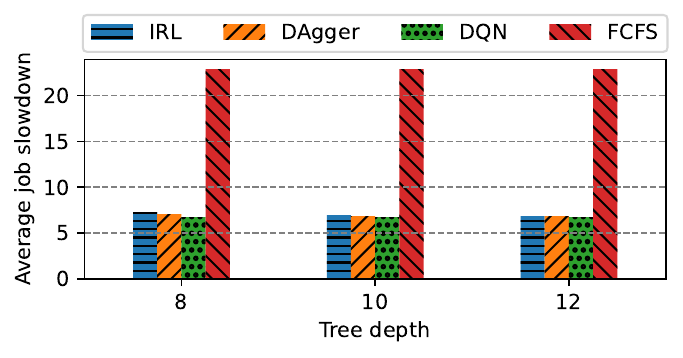}}
    \label{3b}

  \caption{Scheduling performance on SP2 workload trace. The tree depth is set to 8, 10, 12 separately in IRL and DAgger.}
  \vspace{-10pt}
  \label{fig:sp2} 
\end{figure*}

\begin{figure*} 
    \centering
  \subfloat[Average job wait time]{%
       \includegraphics[width=0.40\linewidth]{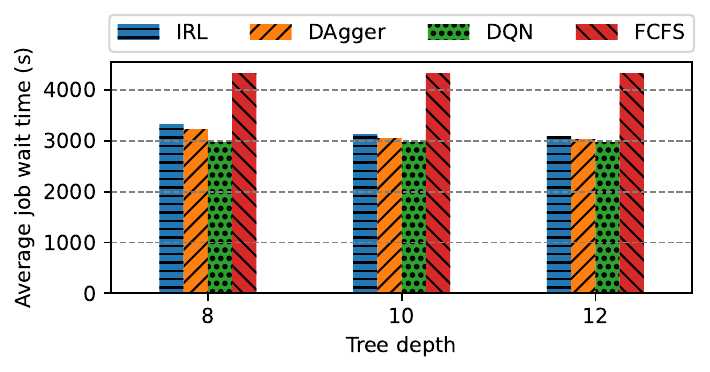}}
    \label{3a}\hspace*{0.06\textwidth}
  \subfloat[Average job slowdown]{%
        \includegraphics[width=0.40\linewidth]{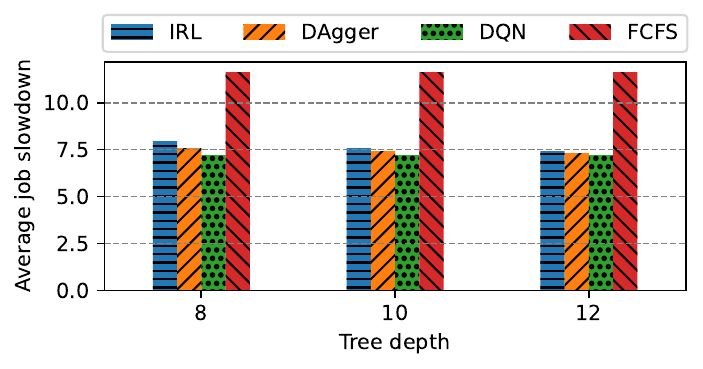}}
    \label{3b}

  \caption{Scheduling performance on DataStar workload trace. The tree depth is set to 8, 10, 12 separately in IRL and DAgger.}
  \vspace{-10pt}
  \label{fig:datastar} 
\end{figure*}

\subsection{Scheduling Performance} \label{sec:scheudling performance}

Figure \ref{fig:sp2} and Figure \ref{fig:datastar} compare different scheduling methods under SP2 and DataStar workloads in terms of average job slowdown and average job wait time. Different tree depths are used in IRL and DAgger to observe the impact of the tree depth on the scheduling performance. 

We observe that IRL yields much better scheduling performance than FCFS. On SP2, IRL can reduce average job wait time and average job slowdown by up to 70\% compared to FCFS. On DataStar, the average job wait time and slowdown can be shortened by up to 36\% in comparison with FCFS. In addition, we notice that IRL achieves comparable scheduling performance in contrast to DQN. On SP2, when the tree depth is 10 or 12, the increase in average job wait time and slowdown of IRL is within 3\% compared to DQN. On DataStar, when the tree depth is 10 or 12, the increase in average job wait time and slowdown of IRL compared to DQN remains within 5\%. 

We also compare the performance between IRL and DAgger to examine the impact of critical state on the scheduling performance. On SP2, when the tree depth is 10 or 12, the scheduling performance loss between IRL and DAgger is within 1\%. On DataStar, when the tree depth is 10 or 12, the increase in average job wait time and slowdown remains within 2\%. This suggests that the introduction of critical state causes negligible performance loss.

\subsection{Tree Reduction} \label{sec:dt_size}

\begin{figure} 
    \centering
  \subfloat[SP2]{%
       \includegraphics[width=0.65\linewidth]{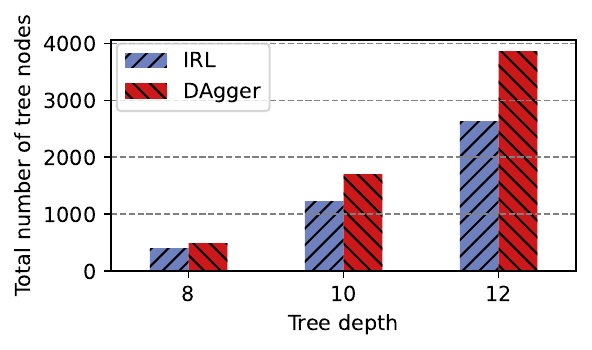}}
    \label{2a}\hfill
  \subfloat[DataStar]{%
        \includegraphics[width=0.65\linewidth]{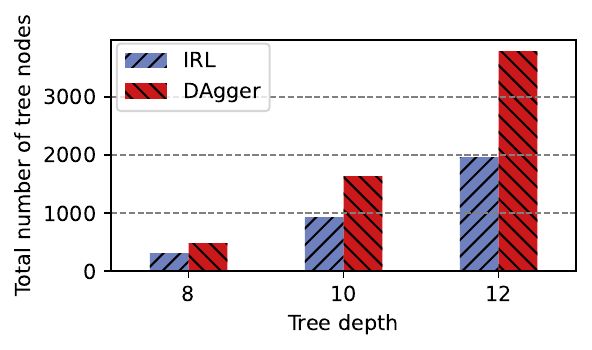}}
  \caption{Comparison of tree size generated by IRL and DAgger. }
  \label{fig:compsize} 
\vspace{-0.45cm}
\end{figure}
The time cost of making decisions with the decision tree is proportional to the size of the tree. Thus, it is preferable to have a decision tree that is compact in size. In this subsection, we analyze the impact of introducing critical state on the size of the tree.

Figure \ref{fig:compsize} compares the sizes of the decision tree generated by IRL and DAgger. On the SP2 workload, IRL reduces the tree size by up to 34\% compared to DAgger. On the DataStar workload, IRL delivers a reduction of up to 48\% in tree size.  The reduction of tree size on the DataStar workload is larger than that on the SP2 workload. We attribute this to the relative idleness of the DataStar system compared to the SP2 system. As demonstrated in Figure \ref{fig:sp2} and Figure \ref{fig:datastar}, the average job wait time of the DataStar workload is significantly less than that of the SP2 workload. Consequently, with the introduction of critical state, IRL stands to benefit more in terms of tree size reduction.

\subsection{Runtime Overhead} \label{sec:overhead}
In our experiments, with the tree depth set to 10, IRL takes approximately 0.0003 seconds for each job selection, whereas DQN requires around 0.02 seconds. Hence, IRL introduces significantly less overhead compared to DQN. All the experiments were conducted on a personal computer configured with an Intel 2 GHz quad-core CPU and 16 GB memory.

\section{Conclusions} \label{conclusion}

%Although nowadays it is an active area to explore the potential of applying deep reinforcement learning methods in HPC cluster scheduling, few of them are deployed on production systems.
While DRL driven scheduling exhibits impressive performance compared to heuristic and optimization methods, there are significant limitations that hinder its practical deployment, particularly the lack of model interpretability. The superior performance of DRL driven scheduling stems from the neural network employed in the design; however, the neural network appears as a black box model to system managers as it is  incomprehensible to debug, deploy, and adjust in practice.

In this work, we have presented IRL, an interpretable model for DRL driven scheduling. IRL converts the black-box neural network employed in DRL driven scheduling to an interpretable decision tree model, which not only can represent a complex decision policy, but also is easy for humans to understand. The design of IRL poses several technical challenges. In this work, we have described the detailed strategies for the design of IRL.  Moreover, we have shown the use of IRL via several case studies. Specifically, one case study demonstrates how IRL can contribute to the DRL scheduling design (e.g., reward setting), and another case study shows that IRL can deliver comparable scheduling performance as compared to the existing scheduling methods including the widely used heuristic method and the DRL scheduling method.

While this study focuses on DQN,  IRL can be applied to other DRL methods as well.  As part of our future work, we will investigate other interpretable models such as the regression model to IRL. To the best of our knowledge, this study represents the first exploration of interpretable reinforcement learning scheduling for HPC. We hope this work will pave the way for further investigations in the field of interpretable reinforcement learning scheduling for HPC.

\section*{Acknowledgment}

This work is supported in part by US National Science
Foundation grants CCF-2109316, CCF-2119294, and U.S. Department of Energy, Office of Science, under contract DE-AC02-06CH11357.

\bibliographystyle{IEEEtran}
\bibliography{bib/cluster.bib}
\end{document}